\documentclass[conference]{IEEEtran}
\ifCLASSINFOpdf
\else
\fi
\usepackage{graphicx}
\graphicspath{{Images/}}
\usepackage[bookmarks=false]{hyperref}
\usepackage{cleveref}
\usepackage{graphicx}

\begin{document}
%
\title{Detecting the Presence of Vehicles and Equipment in SAR Imagery Using Image Texture Features}

\author{\IEEEauthorblockN{Michael Harner}
\IEEEauthorblockA{Lockheed Martin Space\\
King of Prussia, Pennsylvania\\
Michael.J.Harner.Jr@lmco.com}
\and
\IEEEauthorblockN{Austen Groener}
\IEEEauthorblockA{Lockheed Martin Space\\
King of Prussia, Pennsylvania\\
Austen.M.Groener@lmco.com}
\and
\IEEEauthorblockN{Mark Pritt}
\IEEEauthorblockA{Lockheed Martin Space\\
Gaithersburg, Maryland\\
Mark.Pritt@lmco.com}}


%


\maketitle

\begin{abstract}
In this work, we present a methodology for monitoring man-made, construction-like activities in low-resolution SAR imagery.  Our source of data is the European Space Agency's Sentinel-1 satellite which provides global coverage at a 12-day revisit rate.  Despite limitations in resolution, our methodology enables us to monitor activity levels (i.e. presence of vehicles, equipment) of a pre-defined location by analyzing the texture of detected SAR imagery.  Using an exploratory dataset, we trained a support vector machine (SVM), a random binary forest, and a fully-connected neural network for classification.  We use Haralick texture features in the VV and VH polarization channels as the input features to our classifiers.  Each classifier showed promising results in being able to distinguish between two possible types of construction-site activity levels.  This paper documents a case study that is centered around monitoring the construction process for oil and gas fracking wells.

\end{abstract}


%
\IEEEpeerreviewmaketitle

\section{Introduction}

Synthetic aperture radar (SAR) has proven to be a superior imaging modality in many remote sensing applications.  SAR imaging is largely robust to the presence of environmental factors like cloud cover, local weather, and seeing conditions, and due to its active nature, is completely independent of time of day.  This is significant because it was shown in \cite{VanderWal} that, despite the abundance of optical data available, cloud cover can mask up to 75 percent of that data in certain areas of the world.  The reliability SAR provides makes it rather advantageous for monitoring applications that require imagery to be delivered on a consistent basis.      

The intensity of a SAR pixel is directly related to the electrical scattering properties of the area in which the pixel covers.  The scattering properties for a given point scatterer are affected by a multitude of factors including electrical conductivity, surface roughness, shape, size, and it spatial orientation with respect to the incident electro-magnetic wave, to name a few.  A bright SAR pixel is indicative of an area with a very strong radar return.  Generally speaking, a strong return is usually due to an area having high reflectivity and, thus, a high conductivity.  Highly conductive objects tend to be metallic in nature; things like vehicles, buildings, construction equipment.  In addition to this, the geometry of an object will also have a substantial impact on its overall reflectivity.  Objects with sharp edges, similar in shape to a corner reflector, will present very strong radar returns \cite{sandia}. 

SAR techniques have been used for numerous monitoring applications such as crop growth \cite{Steele-Dunne}, ground deformation \cite{Colesanti}, and urban agglomeration \cite{Zhang}.  In many SAR applications, the mean backscatter over a cluster of pixels can be used to classify ground activity at these given pixel locations.  With enough prior knowledge, thresholding techniques can be used to perform this classification.  Although this technique has proven successful for many applications, it is susceptible to the existence of highly reflective point scatterers in the ground plane.  These scatterers can produce strong returns for several reasons: the incidence and aspect angles (collection geometry), and relative size with respect to the radar's transmit frequency.  Due to their strong reflections, they cause the overall scene magnitude to greatly increase.  If the reflections are strong enough, the average magnitude of the scene may surpass a threshold and cause a false classification of the area.  

In \cite{Steele-Dunne}, the authors perform a time series analysis of different crop parcels over the course of a year using normalized radar cross sections (RCS) of the land.  By averaging across hundreds of different parcels, the technique is able to yield a valid assessment of crop growth during this time period.  If the area of interest (AOI) only consisted of 2 or 3 parcels, however, one might imagine how a few strong point scatterers (i.e. a tractor, plow) could skew the overall RCS average for any given moment in time.  In the context of land characterization, this could potentially lead to a misrepresentation of soil moisture for example.

Especially in low-resolution SAR, where each pixel represents a fairly large area of land, there is value in analyzing the spatial components of an image.  This methodology tends to steer away from traditional techniques that average pixel data over an area and disregard where the pixel data exists in space.  Rather, this technique analyzes the spatial distribution of the backscatter for a given scene and learns spatial patterns within the data.

\section{Image Texture}

All surfaces have a unique texture - crop fields, roadways, bodies of water. Each texture represents a surface's structural arrangement and its relationship to the surrounding environment \cite{Haralick}.  Texture is a byproduct of the spatial relationships within an image on a pixel-by-pixel basis. With SAR imagery, and especially at low-resolutions, textural analysis can be utilized for classifying surfaces not so obvious to the human observer.  A single 20-meter resolution cell represents an area of 400 square meters and, thus, the backscattered energy is the summation of reflected energy from all of the smaller objects within the region.  These smaller, individual objects are too small to be resolved by the radar.  Though a cluster of pixels may be uninterpretable by the human observer, the underlying texture of the cluster may be indicative of a particular object or surface.  For this reason, we investigate the feasibility of using image texture in low-resolution SAR imagery as a means for classifying large-scale construction activities.

\subsection{Grey Level Co-Occurrence Matrix}

The grey level co-occurrence matrix (GLCM) is a useful tool used in image processing for analyzing spatial relationships within an image.  For a given image, the GLCM is computed using the following equation:

\begin{equation}
\resizebox{0.91\hsize}{!}{$%
C_{\Delta x, \Delta y} (i, j) = \sum_{x=1}^{n} \sum_{y=1}^{m} \left\{ \begin{array}{ll}
         1, &  \mbox{if } I(x,y) = i \mbox{ and } I(x+\Delta x, y+\Delta y) = j\\
         0, &  \mbox{otherwise}
    \end{array} \right.
$%
}
\end{equation}

where $i$ and $j$ are the pixel values at a given image location $(x,y)$ and $\Delta x, \Delta y$ correspond to a vertical and horizontal pixel-offset within the image.  The values of the computed GLCM matrix, $P_{ij}(s,\theta)$, represent the number of times any two pixels, separated by a distance $s$ and at an angle of $\theta$, have grey-level pixel values of $i$ and $j$, respectively \cite{Liew}.  The main diagonal of a GLCM represents instances where pixel-pairs are equal in value.  Regions near this diagonal represent pixel pairs that are not equal but still close in value.  The regions further away from the diagonal represent pixel-pairs that are significantly different in value.  Haralick texture features can then be derived from the GLCM which resemble various spatial relationships between pixels within the image. 

\subsection{Haralick Texture Features}

Haralick \cite{Haralick} derived multiple statistically-based textural features that analyzed the relative frequency distribution of the grey tones within an image.  Each feature describes a unique spatial pattern within the image.  For our work with low-resolution SAR imagery, we choose to use the following Haralick features: contrast, dissimilararity, homogeneity, angular second moment (ASM), energy, and correlation.  The features, in order, can be derived by the following equations

\hfill

\begin{equation} 
    f_1 = \sum_{i,j=0}^{N} P_{i,j}(i-j)^2\label{eq:2}
\end{equation}
\begin{equation} 
    f_2 = \sum_{i,j=0}^{N} P_{i,j}|i-j|\label{eq:3}
\end{equation}
\begin{equation} 
    f_3 = \sum_{i,j=0}^{N} \frac{P_{i,j}}{1+(i-j)^2}\label{eq:4}
\end{equation}
\begin{equation} 
    f_4 = \sum_{i,j=0}^{N} P_{i,j}^2\label{eq:5}
\end{equation}
\begin{equation} 
    f_5 = \sqrt{ASM}\label{eq:6}
\end{equation}
\begin{equation} 
    f_6 = \sum_{i,j=0}^{N}P_{i,j}\left[\frac{(i-\mu_i)(j-\mu_j)}{\sqrt{(\sigma_i^2)(\sigma_j^2)}}\right]\label{eq:7}
\end{equation}

\hfill

Each texture feature is described in \cite{texture}. Contrast and dissimilarity are both similar in that they are large when the difference in value between two pixels is large.  As shown in Equations \ref{eq:2} and \ref{eq:3}, each value in the summation is calculated by multiplying its corresponding GLCM value with a weight.  For contrast, this weight value increases exponentially as one moves away from the main diagonal.  For dissimilarity, the weight value increases linearly.  As shown in Equation \ref{eq:4}, the weight values for the homogeneity feature decrease exponentially as one moves away from the diagonal.  Because of this, homogeneity tends to be high when the GLCM is mostly focused around the main diagonal (i.e. images with similar pixel values throughout the entire image).  As shown in Equations \ref{eq:5} and \ref{eq:6}, ASM and energy values are calculated by using the GLCM values as weights themselves.  Maximum ASM and energy will occur in an image that has the same value for every pixel.  Correlation, as shown in Equation \ref{eq:7}, measures the relationship between two pixels.  High correlation between two pixels means that there is a predictable and linear relationship between these pixels that can be expressed by a regression equation \cite{texture}.

\section{Data} \label{Data}

The data used in this paper comes from the European Space Agency (ESA) Sentinel-1 satellite constellation.  Sentinel-1's orbit is sun-synchronous with a repeat cycle of twelve days.  The platform offers both co-polarization (VV) and cross-polarization (VH) data channels.  The final Ground Range Detected (GRD) product is an image that has been multi-looked and projected to the ground range. The pixels have a 15 by 15 meter ground sampling distance (GSD) and contain only the amplitude data.

For this work, we have also post-processed the GRD data so that it is $\gamma^0$ calibrated. This calibration step is essential when comparing pixels between multiple images taken in time.  To get the $\gamma^0$ data, one must first generate the $\sigma^0$ data.  To achieve this, the following equation \cite{SARguide} is used:

\begin{equation}
    \sigma^0 = 10*\log_{10}(DN^2) + K
\end{equation}
where $DN$ is the digital number of the pixel and $K$ is the calibration factor for the satellite.  The $\sigma^0$ value is a measure of the overall backscatter return for a given pixel.  This returned backscatter is a summation of all of the smaller, individual scatters that exist within each resolution cell.  In general, $\sigma^0$  is highly dependent on the radar's incidence angle. This phenomenon can cause significant fluctuations in pixel brightness between two geo-registered images taken at different collection geometries.  To account for this, the data can be processed into $\gamma^0$ data which works to mitigate this effect.  Calibration is achieved by
\begin{equation}
    \gamma^0 = \frac{\sigma^0}{\cos(\phi)}
\end{equation}
where $\phi$ is the incidence angle.  In this study, we choose to collect data from the same relative orbit so that effects due to changes in incidence angle are low.  This twelve day collection interval allows us to monitor any location in the world, roughly on a bi-weekly basis.

\section{Methodology}

For a given area of interest, we gather a $\gamma^0$ corrected grey-scale SAR image and calculate the image's GLCM.  This allows us to extract the Haralick features as described above in Equations \ref{eq:2}-\ref{eq:7}.  We do this for both polarization channels which yields us a total of twelve textures.  These texture values are then fed into a classifier which predicts a classification label for the given image.  In this study, we use three different classification methods: random forest, support vector machine (SVM), and a fully-connected neural network.  Each classifier requires a supervised-learning approach and, thus, a collection of labeled truth data for training.

\section{Case Study: Oil and Gas Fracking Wells}

\subsection{Background}
\begin{figure}
    \centering
    \includegraphics[width=8cm, height=6cm]{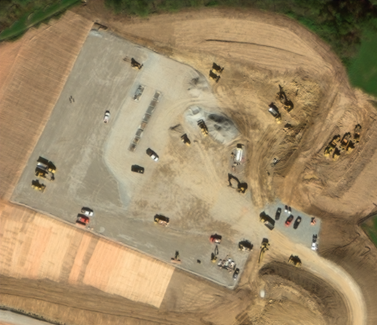}
    \caption{This is an example of a fracking well during the construction phase.  The ground has been excavated and the rock formation is being laid down to form the well pad. Image: DigitalGlobe, Inc.}
    \label{fig:construction}
\end{figure}

\begin{figure}
    \centering
    \includegraphics[width=8cm, height=6cm]{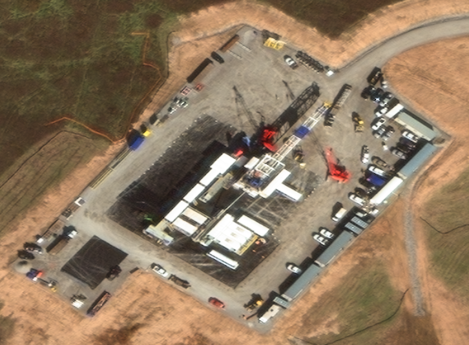}
    \caption{This is an example of a fracking well during the drilling phase.  A large drill tower can be seen in the center of the pad with multiple vehicles and other equipment surrounding. Image: DigitalGlobe, Inc.}
    \label{fig:drilling}
\end{figure}

\begin{figure}
    \centering
    \includegraphics[width=8cm, height=6cm]{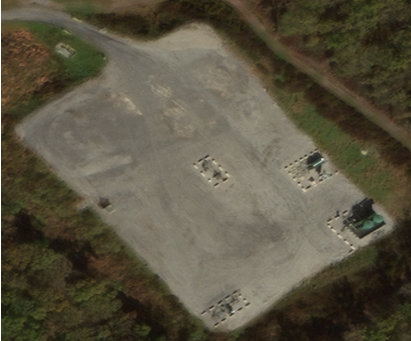}
    \caption{This is an example of a fracking well during the production phase.  The entire fracking process has been completed and the well will remain in this phase for 10-20 years.  The only equipment on the pad are some holding tanks and the well heads. Image: DigitalGlobe, Inc.}
    \label{fig:production}
\end{figure}

Hydraulic fracturing, commonly known as “fracking”, is a process in which natural gas and oil are extracted from natural shale rock buried deep below the Earth’s surface \cite{fracking}.  The process involves drilling a vertical wellbore 1-2 miles beneath the surface and then drilling horizontally for another mile or so along the bottom of the wellbore.  Once the drilling is finished the surrounding rock along the horizontal path is fractured.  Upon fracturing, the natural gas and oil escapes from the shale rock and flows freely up to the surface where it is captured and stored.  The entire process can be broken down into four phases: construction of the pad, drilling, fracturing, and production.
\newline \newline
{\bf Construction:} The construction phase involves lying down a rock foundation to form a large pad that will provide a surface for all of the heavy equipment to rest on.  This process starts in an open field or forested area where the ground must be completely excavated using various types of common construction equipment.  Once this is done, a limestone layer is laid down to form a large 150 meter by 150 meter foundation.  Upon completion of this step, the well pad is formed and the heavy equipment is ready to be loaded on-site.  This step is shown in Figure \ref{fig:construction}.  
\newline \newline
{\bf Drilling:} During the drilling phase, the pad becomes much more active as the equipment is shipped in and loaded on-site (see Figure \ref{fig:drilling}).  The main feature consists of a tall drill tower that is centered on the well pad where it will be used to drill up to 10,000 feet beneath the ground surface.  Along with the drill, there are many other large pieces of equipment that crowd the pad such as storage tanks, storage containers, portable offices, and vehicles.  Once the drilling is completed, steel casings are cemented into the wellbore to provide wall support as well as isolation from the surrounding area. After this phase is complete, the majority of the equipment on the well pad is cleared off and taken away.
\newline \newline
{\bf Hydraulic Fracturing:} During the fracturing phase, the well pad once again becomes heavily populated with equipment.  This time, the main feature of the pad is a collection of large pump-trucks that tightly crowd themselves around the wellbore.  In addition, there are many large storage units loaded onto the pad that hold the necessary materials for making a fluid that will be used during the fracturing process.  The goal of the fracturing phase is to fracture the shale rock that lies along the horizontal portion of the well.  This is done by sending a perforating gun down the wellbore which will use explosives to punch small holes in the casing and create micro-fractures in the shale rock.  After this step, the pump-trucks work together to pump fracking fluids down the wellbore at extremely high pressures.  The sand particles within this fluid flow into the micro-fractures of the shale rock and keep the rock pried open.  After the fracturing process is completed, the well pad is again cleared off.        
\newline \newline
{\bf Production:} The production phase can be classified as all activities that occur post-fracturing.  Up to this point, the wells have been drilled, fractured, and the resources are slowly flowing up through the wellbore where they will be stored in holding tanks before being shipped off or transported along buried pipelines.  The only equipment that remains are a few cylindrical holding tanks and the wellheads themselves.  A fracking well will continue to produce for upwards of twenty years.  Over this time, it is common to see large portions of the well pad foundation removed in an attempt to restore the surrounding land.  This step is shown in Figure \ref{fig:production}.

\subsection{Objective}

The objective of this case study is to classify whether or not a fracking well is in either a drilling or fracturing phase.  We classify these two phases together under the same category of ``Activity''.  We classify all other phases (i.e. construction of the pad, production) as ``Little/No Activity''. Many other types of serendipitous events - like construction monitoring, troop buildup, or large events (like concerts, fairs, and sporting events) can similarly be formulated as an activity classification problems.    

\subsection{Train/Test Dataset}

The data used for training our classifiers consisted of 251 Sentinel-1 images pre-processed as described in Section \ref{Data}.  To form our dataset, we pulled imagery from five different fracking well locations.  At each location we gathered all of the imagery available within the last two years (see Figure \ref{fig:timeseries}).  Due to the lack of precise ground truth for reference, we manually labeled the data based upon our definition of activity (i.e. the drilling and fracking phases). Fortunately, due to the level of activity on the pads, these stages are relatively easy to see within the imagery by eye. Our test set consisted of 125 image chips, from the five fracking wells, selected at random times throughout the fracking process.

\subsection{Training Details}

The random forest model was trained using a thousand trees.  For the SVM, a radial basis function was used as the kernel type.  For the fully-connected neural network, we found that a 12-10-6-4-1 architecture worked best.  The net was trained for a thousand epochs using a binary cross-entropy loss function. To reiterate, our twelve input features are twelve different texture feature values calculated for the given image.  For each of the three classifiers, we trained ten separate models with a different random seed each time.  We then averaged the ten scores together to come up with a reliable accuracy score for each of the three classifiers.

\subsection{Results}

\begin{figure}
    \centering
    \includegraphics[width=8cm, height=7cm]{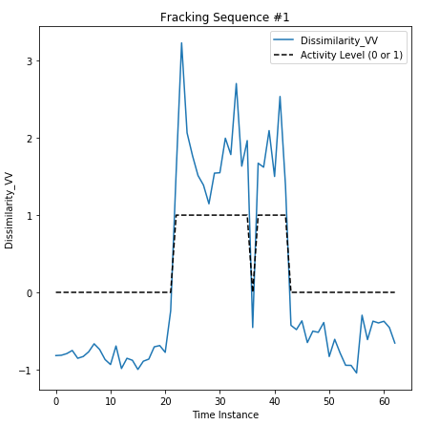}
    \caption{Dissimilarity (VV) values for two years worth of imagery at a given fracking well location.  The values tend to increase during times of high activity and decrease during times of low activity.}
    \label{fig:DISS}
\end{figure}

\begin{figure}
    \centering
    \includegraphics[width=8cm, height=7cm]{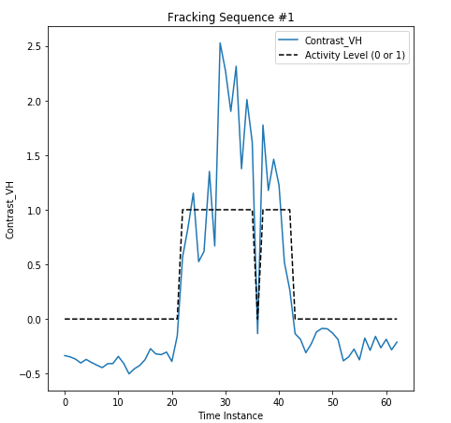}
    \caption{Contrast (VH) values for two years worth of imagery at a given fracking well location.  The values tend to increase during times of high activity and decrease during times of low activity.}
    \label{fig:CONT}
\end{figure}

Overall, all of the models demonstrated successful classification abilities for identifying the activity state of a fracking well.  The random forest, SVM, and fully-connected neural network showed accuracy scores of 95.8, 96.6, and 94.2 percent, respectively, on the same test data.  As shown in Figures \ref{fig:DISS} and \ref{fig:CONT}, certain texture features are high during the drilling and fracturing phases, while remaining small (or negative) during levels of inactivity.  Yet during the other stages of low activity (i.e. construction, production), these values tend to drop off significantly.  In Figures \ref{fig:DISS} and \ref{fig:CONT}, a clear drop in activity occurs at the thirty-seventh time instance in between two instances of high activity.  This corresponds to a stage where the drilling equipment has been removed, and the pad remains cleared until the fracking equipment is loaded on-site.  During this stage, the texture of the image undergoes significant change, thus, alluding to a period of ``No Activity."  

In the case of the random forest classifier, the two most important features were  dissimilarity in the VV channel and contrast in the VH channel.  These two features had importance values of 50 and 26 percent, respectively.  An image with high contrast and dissimilarity will have many values within its GLCM matrix that are concentrated away from the main diagonal.  For contrast, the weights for these off-diagonal values increase exponentially as one moves further away from the diagonal.  Dissimilarity acts similarly, but with linearly increasing weights.  Both correspond to an image pixel pair that differs greatly in value.  For the case of fracking wells, the contrast and dissimilarity values spike during stages of high activity because clusters of bright pixels are introduced into the scene.  This is in contrast to times of low activity where the majority of the pixels within the image are low in value and do not vary much.  The texture features encapsulate the spatial changes within the image very effectively.

\section{Conclusion and Future Work}

In this study, we have shown the potential for using image texture as an avenue for monitoring and/or classification applications.  Unlike some thresholding techniques, this method is robust to scenes with strong point scatterers present.  By analyzing the spatial relationships of an image, on a pixel-by-pixel basis, we can accurately train a classifier using a relatively small dataset.  

We applied these classification capabilities to fracking wells where we were able to produce accuracies in the mid-90's.  This case study is relatable to many other types of large-scale construction activities.  Moving forward, we would like to investigate this technique on a larger and more diverse dataset.  We would also like to investigate using this technique to perform multi-class classification on other types of construction sites.  This could  be used in many other applications where the area on the ground is undergoing some type of change.  Some potential use-cases include agriculture for crop monitoring, traffic analysis, and/or flooding assessment for disaster-relief applications.


\section*{Acknowledgment}

The authors would like to thank DigitalGlobe, Inc. and Descartes Labs, Inc. for providing all of the imagery used for this study.

\vfill\null



%


\end{document}